\documentclass[10pt,twocolumn,letterpaper]{article}

\usepackage[pagenumbers]{cvpr} % To force page numbers, e.g. for an arXiv version
\usepackage{multirow}
\usepackage{graphicx}
\usepackage{tikz}
\usepackage{multicol}
\usepackage[misc]{ifsym}
\definecolor{cvprblue}{rgb}{0.21,0.49,0.74}
\usepackage[pagebackref,breaklinks,colorlinks,citecolor=cvprblue]{hyperref}

\newcommand{\figref}[1]{Fig.~\ref{#1}}
\newcommand{\tabref}[1]{Tab.~\ref{#1}}

\def\paperID{3189} % *** Enter the Paper ID here
\def\confName{CVPR}
\def\confYear{2024}

\makeatletter
\def\thanks#1{\protected@xdef\@thanks{\@thanks
        \protect\footnotetext{#1}}}
\makeatother

\title{Decoupling Degradation and Content Processing for Adverse Weather \\ Image Restoration}

\begin{document}

\author{Xi Wang\textsuperscript{\rm1,2 $^{\star}$},
    Xueyang Fu\textsuperscript{1 $^{\textrm{\Letter}}$},
	  Peng-Tao Jiang\textsuperscript{\rm2 $^{\textrm{\Letter}}$}, 
	  Jie Huang\textsuperscript{\rm2}, \\
	  Mi Zhou\textsuperscript{\rm2},
    Bo Li\textsuperscript{\rm2},
    Zheng-Jun Zha\textsuperscript{\rm 1} \\
\textsuperscript{\rm 1} University of Science and Technology of China \quad 
 \textsuperscript{\rm 2} vivo Mobile Communication Co., Ltd  \\ 
{\tt\small  wangxxi@mail.ustc.edu.cn, xyfu@ustc.edu.cn, pt.jiang@vivo.com
}
}
\thanks{ 
\hspace{-6mm}  ${\star}$ : This work was done during the internship at 
vivo Mobile Communication Co., Ltd.  Project was led by Peng-Tao Jiang. \\
 ${\textrm{\Letter}}$ : Corresponding authors. }

\maketitle
\begin{abstract}
Adverse weather image restoration strives to recover clear images from those affected by various weather types, such as rain, haze, and snow. Each weather type calls for a tailored degradation removal approach due to its unique impact on images. Conversely, content reconstruction can employ a uniform approach, as the underlying image content remains consistent. Although previous techniques can handle multiple weather types within a single network, they neglect the crucial distinction between these two processes, limiting the quality of restored images. 

This work introduces a novel adverse weather image restoration method, called DDCNet, which decouples the degradation removal and content reconstruction process at the feature level based on their channel statistics. Specifically, we exploit the unique advantages of the Fourier transform in both these two processes: (1) the degradation information is mainly located in the amplitude component of the Fourier domain, and (2) the Fourier domain contains global information. The former facilitates channel-dependent degradation removal operation, allowing the network to tailor responses to various adverse weather types; the latter, by integrating Fourier's global properties into channel-independent content features, enhances network capacity for consistent global content reconstruction. We further augment the degradation removal process with a degradation mapping loss function. Extensive experiments demonstrate our method achieves state-of-the-art performance in multiple adverse weather removal benchmarks.

\end{abstract}    
\section{Introduction}
\label{sec:intro}

\begin{figure}[t]

\centering

\includegraphics[width=1.0\linewidth]{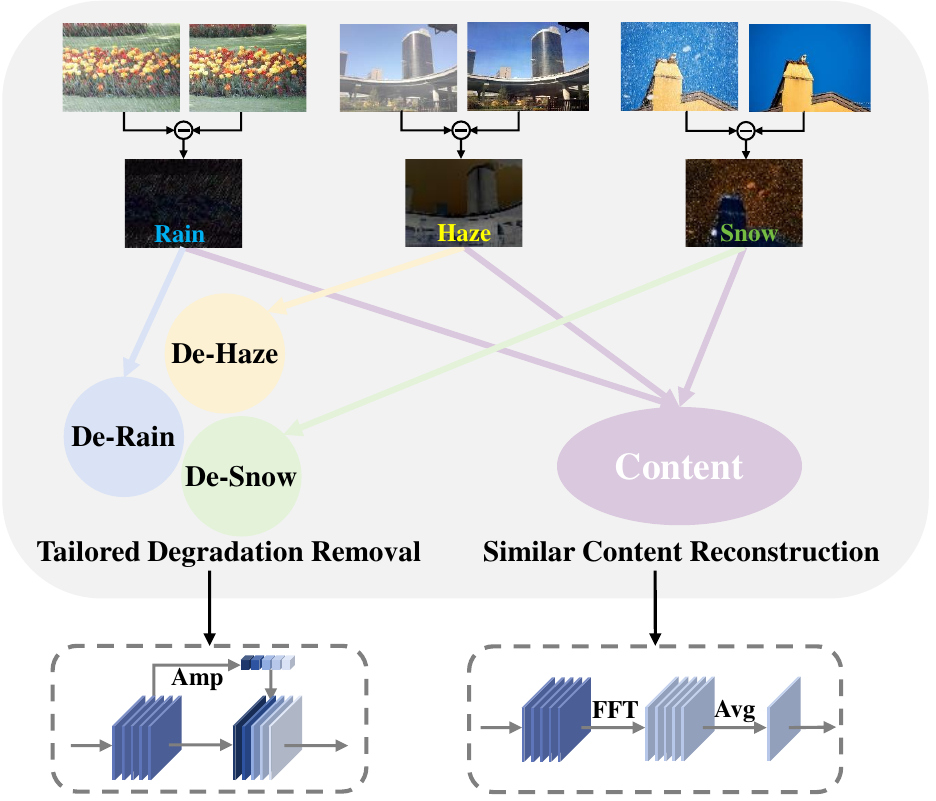} 
% \fbox{\rule{0pt}{3in} \rule{.9\linewidth}{0pt}}
% \vspace{0pt}
\caption{Adverse weather image restoration requires degradation removal and content reconstruction simultaneously.
Different weather types necessitate tailored degradation removal due to their unique effects on images 
while the content reconstruction could be similar as the underlying image content remains consistent.
In this paper, we decouple these two processes at the feature level based on their channel statistics and exploit the unique advantages of the Fourier transform to accomplish it. 
% (1) The degradation information is primarily present in the amplitude component of the Fourier domain, 
% and (2) the Fourier domain encompasses global information. 
% These characteristics allow us to achieve tailored channel-dependent degradation removal and consistent channel-independent global content reconstruction.
} 
\vspace{-2pt}

\label{fig:teaser}
\end{figure}

Adverse weather conditions such as rain, haze, and snow commonly affect images,  not only diminishing their visual quality but also hindering computer vision tasks like autonomous driving and object detection. To restore these adverse weather images, numerous algorithms have been developed over time. Initially, researchers employ handcraft-designed model priors \cite{zhu2017joint_hand, liu2019removing_hand} to address various weather conditions. However, the inflexibility of these handcrafted model priors frequently resulted in less effective restoration outcomes, failing to adequately address the complexities of varying weather impacts. 
With the rapid advancement of deep learning, learning-based methods \cite{fu2017removing_rain_rain1400, wang2021multi_rains, ren2016single_haze,  liu2018desnownet_snow100k} have exhibited remarkable abilities, many deep learning methods have emerged for adverse weather restoration. Early approaches \cite{fu2017clearing_rain, he2010single_haze,chen2020jstasr_snow} focus on employing a single model to tackle one specific type of adverse weather condition. However, the performance of these networks is limited when faced with weather types they are not specifically designed for,  limiting their usage flexibility.  Hence, employing a single network to manage various adverse weather conditions is essential and holds substantial research importance. 

Despite the growing number of methods \cite{chen2022learning_all, li2020all_all-in-one_all, valanarasu2022transweather_all, zhu2023learning_all, ozdenizci2023restoring_diffusion_all} proposed to handle different degraded weather images using a unified framework, they all have certain shortcomings. All-in-One \cite{li2020all_all-in-one_all} introduces a network capable of restoring images under a variety of adverse weather conditions, however, its reliance on multiple encoders makes it computationally intensive. 
Transweather \cite{valanarasu2022transweather_all} integrates transformer architecture for the effective removal of severe weather effects, utilizing cross-attention and weather-type queries in the decoder for a comprehensive approach to adverse weather removal.  
However, it fits degradation information from the data,  without considering the inherent degradation characteristics in the image itself.  Zhu \textit{et al.} \cite{zhu2023learning_all} use a two-stage strategy, considering the general and specific characteristics of different adverse weather conditions separately. However, this model requires knowledge of weather types during inference, limiting its flexibility. 

In fact, all previous methods have overlooked the necessity of separating the degradation removal and content reconstruction process for different weather types. Different weather types should be tailored for degradation removal due to their unique effects on images, while the content reconstruction can be similar as the underlying image content remains consistent for each weather type. 
Moreover, previous methods do not sufficiently mine the degradation information inherent in the image itself.

To address the above challenges, this paper introduces a novel approach, called DDCNet, 
to recover images of various adverse weather types in a unified network. As shown in \figref{fig:teaser}, we decouple the degradation removal and the content reconstruction process into distinct modules at the feature level based on their channel statistics. 
Due to the variation in channel distribution among different degradation types, we modulate the features along the channel dimension to achieve channel-dependence degradation removal. 
Moreover, recognizing the crucial role of global information in content reconstruction, we employ it to facilitate the attainment of consistent image reconstruction. 
Owing to the rich information brought by the Fourier transform, we utilize it to assist 
these two processes.
% Owing to the exclusive advantages of \jpt{the information brought by the Fourier transform}, we utilize Fourier to assist these two processes.
%
Specifically, in the degradation removal module, we utilize the amplitude of the Fourier domain, which primarily contains image degradation information, extracting channel-dependence degradation information to adjust the importance between channels. This allows the network to generate tailored responses for various degradations.
In the content reconstruction module, we extract the channel-independence content feature and utilize the global characteristics of Fourier space, allowing the network to process the background content of any weather condition in a consistent and global manner.  
Additionally, we integrate a degradation mapping loss function to constrain the direction of degradation removal.

% Our method simplifies network complexities by dividing the degradation removal and content reconstruction process. Furthermore, it mines degradation characteristics in the image itself, removing the requirement for prior knowledge of weather types, thereby improving network efficiency and usability.
%
The contributions of this paper could be summarized as:

\begin{itemize}
\setlength{\leftskip}{0.5cm}
\setlength{\rightskip}{0cm} 
\item This research first highlights the tailored degradation removal and similar content reconstruction processes for various adverse weather types. By decoupling these two processes based on their channel statistics, it reduces the complexity of the unified network in handling multiple adverse weather types. 

\end{itemize}

\begin{itemize}
\setlength{\leftskip}{0.5cm} 
\setlength{\rightskip}{0cm}
\item We propose a new method for channel-dependent degradation removal by utilizing the degradation-related amplitude part in the Fourier space as an auxiliary component.  Furthermore, we utilize Fourier's global properties to bolster channel-independent globally consistent content reconstruction, enhancing the restoration of weather-impacted images.

\end{itemize}

\begin{itemize}
\setlength{\leftskip}{0.5cm} 
\setlength{\rightskip}{0cm} 
\item The proposed method can restore images of different weather types in a unified 
network. We conduct extensive experiments on several adverse weather removal 
benchmarks and outperform previous multi-task state-of-the-art methods. 
% We propose a degradation optimization direction alignment loss function that incorporates weather-impacted images to further strengthen the degradation removal and constrain the network's optimization process for different weather conditions.

\end{itemize}

\section{Related Works}

Adverse weather removal, such as draining \cite{hu2019depth_rain, wang2020model_rain, fu2017clearing_rain}, raindrop removal \cite{quan2019deep_raind, zhang2021dual_raind, qian2018attentive_raind_raindrop_attngan}, dehazing \cite{he2010single_haze, yang2018towards_haze, ren2016single_haze}, and desnowing \cite{chen2020jstasr_snow, liu2018desnownet_snow100k, zhang2021deep_snow_ddmsnet}, has gained attention in recent years, and we summarize them as follows.

%-------------------------------------------------------------------------
\paragraph{Single Image Deraining.} 
Currently, single image deraining primarily uses CNN-based methods. Fu \textit{et al.} \cite{fu2017clearing_rain}introduced global residual learning into single image deraining and demonstrated the effectiveness of a deeper structure in rain streak removal.  Subsequently, numerous CNN-based rain streak removal methods \cite{wang2021multi_rains, yang2019joint_rains, chen2021robust_rains, zhu2020learning_rains, chen2022unpaired_rains, ye2022unsupervised_rains, zhang2021multifocal_rains, jiang2020multi_rains_mspen, li2019heavy_rains_outdoor_rain}, were proposed, and raindrop removal \cite{quan2019deep_raind, zhang2021dual_raind, hao2019learning_raind, quan2021removing_raind_ccn, qian2018attentive_raind_raindrop_attngan}, which differs from rain streak removal, was also introduced. In order to better restore texture details and improve visual quality, Zhang \textit{et al.} \cite{zhang2019image_rain_gan} utilized GANs to handle rain raindrops. IDT \cite{xiao2022image} combines the advantages of transformers and CNNs, uses non-local information and retains high-frequency components, achieving better results in rain image restoration.

\paragraph{Single Image Dehazing.} 
He \textit{et al.} \cite{he2010single_haze} utilize dark channel prior for Single image haze removal. Subsequently, some methods used CNNs to estimate the parameters of degraded haze images. Some methods \cite{he2010single_haze, ren2016single_haze, yang2018towards_haze, cai2016dehazenet_haze, yang2018proximal_haze, qu2019enhanced_haze_epdn} also directly use an end-to-end learning approach to map haze images directly to clean images. Some work is also dedicated to the restoration of real-world haze images, and it is combined with unsupervised methods. 

\paragraph{Single Image Desnowing.} 
DesnowNet \cite{liu2018desnownet_snow100k} uses a two-stage network learning approach for snow removal. To cope with the challenges posed by snow of multiple sizes and non-transparency, Chen \textit{et al.} 
\cite{chen2020jstasr_snow} proposed a unified approach dealing with size and transparency concurrently, it also achieved the removal of both snow and haze. 

\paragraph{All-in-One Adverse Weather Removal.} 
Different from the removal of a single adverse weather condition, for practicality, an increasing number of studies are focusing on handling multiple weather deteriorations using a single network. Li \textit{et al.} \cite{li2020all_all-in-one_all} use task-specific encoders for each type of adverse weather and use a neural architecture search to optimize the processing of image features extracted from all encoders, thus achieving all-in-one image restoration. For the first time, TransWeather \cite{valanarasu2022transweather_all} introduces the transformer into all-in-one adverse weather restoration. It leverages the attention mechanism in the transformer and sets the query in the decoder as adaptively learned parameters to realize adaptive processing for different weather conditions. Chen \textit{et al.}~\cite{chen2022learning_all} employ a two-stage training strategy, by distilling the knowledge learned from multiple "teacher" networks responsible for different weather conditions in the first stage to the "student" network in the second stage, thus achieving promising results in various bad weather removal tasks. With the popularity of generative models, WeatherDiffusion \cite{ozdenizci2023restoring_diffusion_all} first introduces the diffusion model into adverse weather removal, its patch-based methodology enables the model to recover images of any size while maintaining quality. Zhu \textit{et al.} \cite{zhu2023learning_all} also adopt a two-stage learning strategy, by adaptively adding parameters for different weather types in the second stage to handle various weather conditions.
However, these methods require the introduction of additional network structures or need to know the weather type during inference. Our method considers unique degradation removal and similar content reconstruction for different adverse weather conditions,  reducing the complexity of the unified network in handling multiple adverse weather types.  Furthermore, we achieve blind weather image restoration by exploiting the inherent degradation information present in the image itself.

\begin{figure}[t]

\centering

\includegraphics[width=1.0\linewidth]{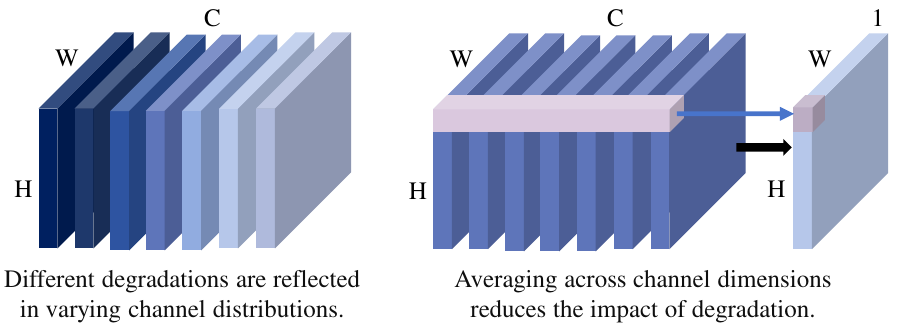} 
% \fbox{\rule{0pt}{3in} \rule{.9\linewidth}{0pt}}
% \vspace{0pt}
\caption{Channel statistics for degradation removal and content reconstruction decoupling.} 
\vspace{-3pt}

\label{fig:decon}
\end{figure}
\section{Method}
\label{sec:formatting}

In this chapter, we start by discussing the necessity of separately handling degradation removal and content reconstruction processes, outlining the motivation behind our proposal. Following this, we detail how to effectively separate these processes and explain the use of the Fourier transform as an auxiliary to assist these processes. Additionally, we introduce the degradation mapping loss function proposed in our approach.

%-------------------------------------------------------------------------
\begin{figure*}[t]

\centering

\includegraphics[width=1.0\linewidth]{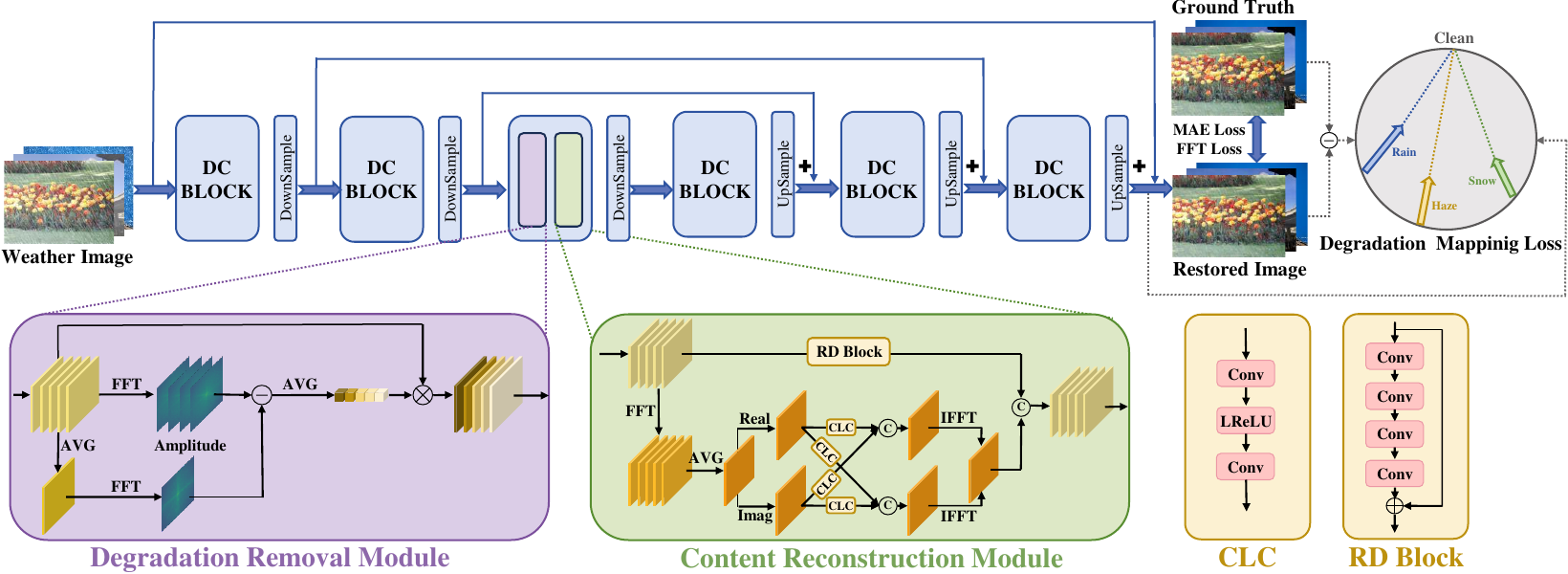} 
% \fbox{\rule{0pt}{3in} \rule{.9\linewidth}{0pt}}
\vspace{-10pt}
\caption{The overall of our proposed UDCNet framework, which manages the degradation removal and content reconstruction in separate modules based on their channel statistics.
Specifically, to achieve tailored channel-dependent degradation removal, we utilize the degradation-related amplitude part in the Fourier domain to modulate the feature along the channel dimension. 
For channel-independent content reconstruction, we average the feature along the channel to eliminate degradation information and use Fourier's global properties to accomplish globally consistent content reconstruction.
Additionally, we introduce a novel degradation mapping loss function, utilizing degraded images to strengthen the degradation removal process.} 
\vspace{0pt}

\label{fig:framework}
\end{figure*}

\subsection{Background and Motivation}

Adverse weather image restoration seeks to recover clear images from those affected by various weather conditions (e.g., rain, haze, and snow). Different weather conditions call for specific processes for degradation removal because they affect images in unique ways. Yet, content reconstruction can follow a single process since the underlying content of the image remains the same. Therefore, separating the processes of degradation removal and content reconstruction can reduce the complexity faced by a single model in handling multiple types of adverse weather conditions. 
While previous techniques address multiple weather conditions within a single network, they have neglected the importance of using specific attributes for these two processes, consequently resulting in sub-optimal results.

In this paper, we decouple the degradation removal and content reconstruction based on their channel statistics, which is shown in \figref{fig:decon}. 
% For degradation removal, adjustments are made to different channels due to the differences in degradation distribution reflected by each channel. 
For degradation removal, we adjust the features along the channel dimension based on the differences in degradation distribution across different weather conditions.
For content reconstruction, the content exists in a channel-independent space, and global information plays a vital role. 
Therefore, we use channel-dependent degradation guidance information and channel-independent global content information to implement these two processes respectively.
To achieve this, we utilize the unique advantages of Fourier to assist these two processes: (1) the degradation information of images is primarily concentrated in the amplitude, and (2) Fourier has the capability to capture global information. Therefore, in the degradation removal module (DRM), we utilize the amplitude to generate distinctive conditional information to guide the network to generate different responses for various types of weather. In the content reconstruction module (CRM), we leverage the global information capture property of Fourier to achieve global content reconstruction. 
Additionally, since there are discernible differences in the optimization directions among various adverse weather conditions, which primarily concentrate on the degradation information, we have further proposed a degradation mapping loss function to constrain their optimization direction.
In this way, the network reduces the difficulty of multiple degradation processing, thereby achieving high-quality restoration results.

%-------------------------------------------------------------------------
\subsection{The Network Framework}
Our objective is to recover clean background contents from images damaged by various adverse weather conditions. 
To achieve our goal, we propose a novel unified framework, called DDCNet. 
Our framework utilizes a multi-scale U-Net architecture that incorporates three encoders and three decoders.  Each encoder and decoder mainly consists of two components: the degradation removal module and the content reconstruction module. This setup includes three upsampling and downsampling operations. 
The input and output dimensions of each module remain consistent.
Our detailed framework is shown in \figref{fig:framework}.
In the following, we will introduce these two important modules in detail.

\paragraph{Degradation Removal Module (DRM)}
In the degradation removal module, as the distribution of different degradations varies across channels, the goal of this module is to modulate feature channels according to the degradation.
% our aim is to modulate the tailored channel-related degeneration removal operation. 
To achieve this, we leverage the observation that degradation information in an image is mainly concentrated in the amplitude part of the Fourier domain. 
By utilizing the amplitude's characteristics, we generate channel-dependent degradation conditional information, enabling the network to focus more on removing adverse weather degradation.

Formally, let $\mathbf{X} \in \mathbb{R}^{H \times W \times C}$ denote the input feature of DRM. 
When inputting $\mathbf{X}$ to DRM, we first apply the Fourier transform to it 
in terms of spatial dimensions, which can be expressed as:
\begin{equation}
  \mathbf{\hat{X}}_{u,v} = \frac{1}{\sqrt{HW}} \sum\limits_{h=0}^{H-1}\sum\limits_{w=0}^{W-1} \mathbf{X}_{h,w} \times e^{-j2\pi(\frac{h}{\mathrm{H}}u + \frac{w}{\mathrm{W}}v)},
  \label{eq:fft}
\end{equation}
where $u, v$ denotes the coordinates of the Fourier space feature $\mathbf{\hat{X}}$.
Then we can generate the amplitude component $\mathbf{A}$ by:
\begin{equation}
  \mathbf{A} = \sqrt{ \mathbf{R}^2 + \mathbf{I}^2 },
  % \mathcal{P}(x)(u, v) = \arctan\left[\frac{\mathcal{I}(x)(u, v)}{\mathcal{R}(x)(u, v)}\right],
  \label{eq:amp}
\end{equation}
where $\mathbf{R}$ and $\mathbf{I}$ are real and imaginary parts of $\mathbf{\hat{X}}$, 
respectively.
% As the distribution of different degradations varies across channels, we aim to extract channel-independent information about degradation conditions.
%
The amplitude $\mathbf{A}$ contains both channel-dependent and channel-independent degradation information, but we only need channel-dependent 
degradation information to modulate feature channels for various weather conditions.
Thus, to generate the channel-dependent degradation information from $\mathbf{A}_{deg}$, we further generate the channel-independent degradation information and remove it from $\mathbf{A}$.
Specifically, we first average the input features $\mathbf{X}$ across the channel dimension and generate the channel-independent feature $\mathbf{X}_{ci} 
\in \mathbb{R}^{H \times W \times 1}$. 
Then, we perform a Fourier transform to $\mathbf{X}_{ci}$ and generate 
its amplitude $\mathbf{A}_{ci}$, where $\mathbf{A}_{ci}$ 
contains channel-independent degradation information.
To extract the differences in degradations across channels, 
we remove $\mathbf{A}_{ci}$ from $\mathbf{A}$:
\begin{equation}
    \mathbf{A}_{deg} = \mathbf{A} - \mathbf{A}_{ci}.
\end{equation}
By the above operations, we can obtain the channel-dependent degradation 
information $\mathbf{A}_{deg}$.
%
% After obtaining the degradation-focused amplitude $\mathbf{A}_{deg}$, 
After that, we transform $\mathbf{A}_{deg}$ to channel-dependent degradation condition $\mathbf{v}_{con} \in \mathbb{R}^{1 \times 1 \times C}$ 
by a project network consisting of a global average pooling layer, 
two consecutive linear layers and a sigmoid function, which is formulated as 
\begin{equation}
    \mathbf{S}_{con} = \mbox{Sigmoid}(f_{p2}(\mbox{ReLU}(f_{p1}(\mathbf{v}_{con})))),
\end{equation}
where $f_{p1}$ and $f_{p2}$ denote the linear layers. 
The condition vector $\mathbf{S}_{con}$ modulates the original features $\mathbf{X}$, 
directing the network involves combining channel information based on important relationships to perform degradation removal operations. 
Consequently, the modulated degradation removal feature $\mathbf{X}_{d} \in \mathbb{R}^{H \times W \times C}$ 
is generated by 
\begin{equation}
    \mathbf{X}_{d} = \mathbf{X} \cdot \mathbf{S}_{con},
\end{equation}
% The modulated feature $\mathbf{X}_{deg}$
where $\mathbf{S}_{con}$ is broadcast to the shape of $\mathbf{X}$ for a dot product.

%-------------------------------------------------------------------------
\paragraph{Content Reconstruction Module (CRM)}
For content reconstruction, both local and global content information plays 
crucial roles. Therefore, in this module, we utilize two branches, one for 
local detail captured and the other for global content captured.
% both local and global branches.

In the global branch,  we pursue processing global content information 
that is independent of degradation.  
Leveraging the power of the Fourier transform to capture global information, our approach begins by converting the degradation removal feature $\mathbf{X}_{deg}$ into the Fourier domain 
as $\mathbf{\mathbf{\hat{X}}}_{deg}$. 
This transformation enables us to encode the spatial-global information of the feature.
Subsequently, drawing inspiration from \cite{li2019positional}, we employ channel dimension averaging of features to minimize the presence of degradation information. By averaging $\mathbf{\hat{X}}_{deg}$ along the channel dimension, we obtain a single-channel feature denoted as $\mathbf{\hat{X}}_{s} \in \mathbb{R}^{H \times W \times 1}$, which represents channel-independent global content information.
%
% Notably, this transformed feature predominantly contains valuable global content information while minimizing the presence of degradation-related details. 
%
Therefore, our model can effectively extract essential information for content reconstruction.
Moving forward, we extract the real part, $\mathbf{\hat{X}}_{real}$, and the imaginary part, $\mathbf{\hat{X}}_{imag}$, from the transformed feature $\mathbf{\hat{X}}_s$. 
%
% To further enhance the discriminative capabilities of these components, 
% we resblock each of them individually.
% %
% This operation serves the purpose of extracting relevant features that complement the content information encoded in each part.
Finally, we synergistically combine the extracted features from the real 
and imaginary branches. 
This integration is accomplished through an interplay of feature interactions, 
which is formulated as follows:
\begin{equation}
\begin{aligned}
\mathbf{X}_{real} &= f_{1} \big( [f_{clc, 11}(\mathbf{X}_{real}), f_{clc,12}(\mathbf{X}_{imag})] \big), \\
\mathbf{\hat{X}}_{imag} &= f_{2} \big( [f_{clc,21}(\mathbf{X}_{real}), f_{clc,22}.(\mathbf{X}_{imag})] \big),
\end{aligned}
\end{equation}
where \verb|[|,\verb|]| represents the concatenation operation, $f_1$ 
and $f_2$  both indicate a 3$\times$3 convolution.
$f_{clc, *}$ indicates a $\mbox{CLC}$ block including two 3$\times$3 convolutions with a LeakyReLU activation function in between. 
Then, we transform the features to the original space $\mathbf{X}_{tem} \in \mathbb{R}^{H \times W \times C}$ by performing IFFT transformations on $\mathbf{\hat{X}}_{real}$ and $\mathbf{\hat{X}}_{imag}$. 

In the global branch, taking the average over the channel dimension inevitably results in some content detail loss. 
% Moreover, the global branch cannot capture details effectively. 
Therefore, in the local branch, we introduce a residual dense block (RDB) consisting of convolutional layers to capture local detailed features.
The RDB helps preserve information and enhance feature representation. 
The input of the RDB is $\mathbf{X}_{deg}$ and the output of the RDB is denoted as $\mathbf{X}_{r} \in \mathbb{R}^{H \times W \times C}$.
Finally, to aggregate global and local information, we concatenate $\mathbf{X}_{tmp}$ and $\mathbf{X}_{r}$ toward the channel dimension. 
The concatenated result is then passed through a 1 $\times$ 1 convolution. 
The final output of this operation is referred to as $\mathbf{X}_{con} \in \mathbb{R}^{H \times W \times C}$.
%-------------------------------------------------------------------------

\subsection{Loss Function}

In this paper, we employ three loss functions to optimize our model, in which 
the proposed degradation mapping loss is specifically designed for handling 
multiple adverse weather conditions. 
Let $\mathbf{I}_{in}$ denote the input adverse weather image, $\mathbf{I}_{out}$ denote 
the network's output image, and $\mathbf{I}_{gt}$ denote the corresponding ground truth.

\paragraph{Degradation Mapping Loss (DM).}
We constrain the network optimization process by incorporating weather type into the loss function. 
%
% Specifically, we maximize the cosine similarity between the input image and the network's output image, 
% as well as between the input image and the ground truth, to enforce consistency in the direction 
% of degradation removal, which is formulated as:
%
Specifically, we first compute the residual between the input image and the network's output image, 
as well as the residual between the input image and the ground truth, then maximize the cosine similarity 
between these residuals to enforce consistency in the direction of degradation removal, 
which is formulated as:
\begin{equation}
  L_{DA}(\mathbf{I}_{in}, \mathbf{I}_{out}, \mathbf{I}_{gt}) = 1 - \left< \mathbf{I}_{out} - \mathbf{I}_{in}, \mathbf{I}_{gt} - \mathbf{I}_{in} \right>,
  \label{eq:dm}
\end{equation}
where $\left<,\right>$ represents the cosine similarity.
%
% \begin{equation}
%   \left< \mathbf{S}_{out}, \mathbf{S}_{gt} \right> = \frac{{\mathbf{S}_{out} \cdot \mathbf{S}_{gt}}}{{\left\| \mathbf{S}_{out} \right\| \times \left\| \mathbf{S}_{gt} \right\|}}.
%   \label{eq:amp}
% \end{equation}

\paragraph{Mean Absolute Error Loss (MAE).} 
Since MAE is insensitive to outliers and has better robustness than mean squared error (MSE) for better edges and textures restoration, we choose MAE to train our network:

\begin{equation}
  L_{MAE}(\mathbf{I}_{out}, \mathbf{I}_{gt}) =  \left\| \mathbf{I}_{gt} - \mathbf{I}_{out} \right\|_{1}.
  \label{eq:mae}
\end{equation}

\paragraph{FFT Loss.} 
Since the frequency domain distribution of degraded images differs from that of clean compartments, we also impose the constraint in the frequency domain.
Specifically, we first conduct the Fast Fourier Transform (FFT) of the resorted image and ground truth, 
then measure the mean absolute error between them, which is formulated as:
\begin{equation}
  L_{FFT}(\mathbf{I}_{out}, \mathbf{I}_{gt}) = \Vert FFT(\mathbf{I}_{gt}) - FFT(\mathbf{I}_{out})\Vert_{1},
  \label{eq:fft_loss}
\end{equation}
where $\mathrm{FFT}$ stands for fast Fourier transform, which converts an image 
to the frequency domain.
Therefore, the overall loss function can be formulated as:
\begin{equation}
  L_{TOTAL} = \lambda_1 L_{MAE} + \lambda_2 L_{FFT} + \lambda_3 L_{DM}.
  \end{equation}

%-------------------------------------------------------------------------
\begin{figure}[t]

\centering

\includegraphics[width=1.0\linewidth]{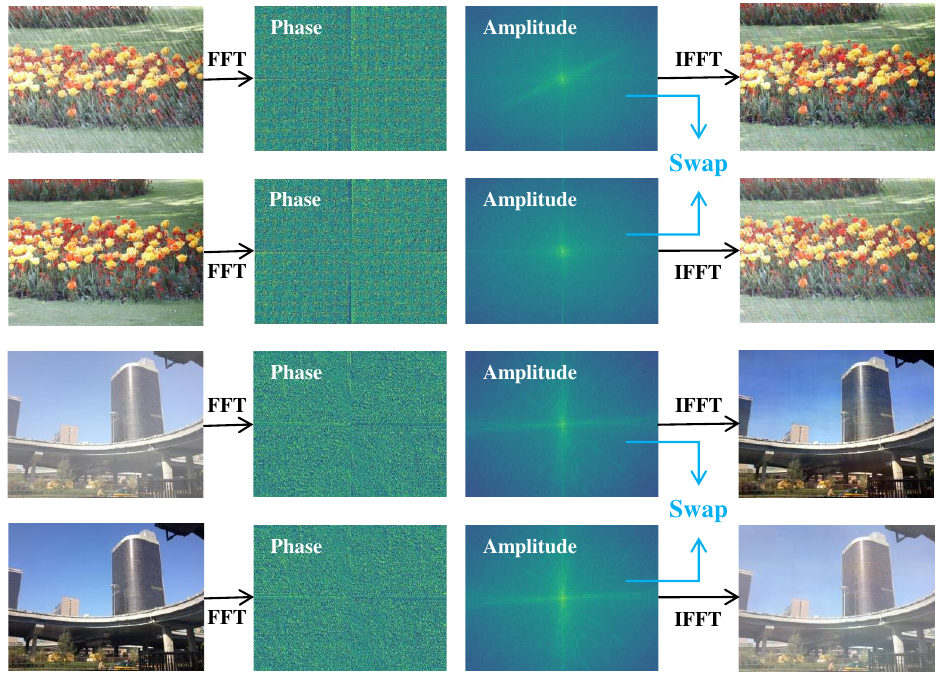} 
% \fbox{\rule{0pt}{3in} \rule{.9\linewidth}{0pt}}
\vspace{-20pt}
\caption{We swap the amplitude of the adverse weather image with its clean compartment, 
as can be seen, the attributes of `degradation' and `clean' are transferred along with the amplitude.} 
\vspace{-8pt}

\label{fig:swap}
\end{figure}

\subsection{Discussion}

\begin{figure}[t]

\centering

\includegraphics[width=1.0\linewidth]{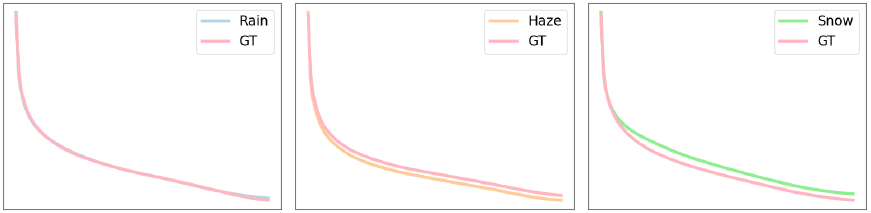} 
% \fbox{\rule{0pt}{3in} \rule{.9\linewidth}{0pt}}
\begin{tikzpicture}[overlay, remember picture]
  \node at (0, 0)[xshift=-0.33\linewidth] {(a)};  % 根据需要调整位置
  \node at (0, 0) {(b)};
  \node at (0, 0)[xshift=0.33\linewidth] {(c)};
\end{tikzpicture}
\vspace{0pt}
\caption{We randomly select 500 images from each adverse weather dataset in \textbf{[Setting2]}, 
and plot the amplitude statistics of these images, labeled as (a), (b), and (c) for rain, haze, and snow, respectively. 
These graphs distinctly illustrate that different adverse weather types exhibit unique optimization directions in their frequency amplitude.} 
\vspace{0pt}

\label{fig:mag}
\end{figure}
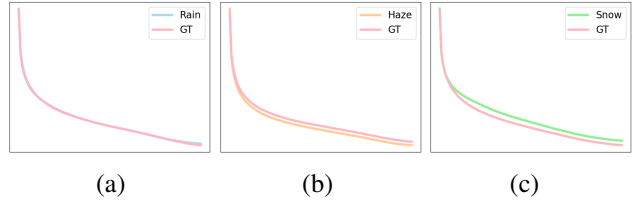

To further understand the rationality of using Fourier transforms for degradation removal and content reconstruction process, we conduct some visual analysis as demonstrated in \figref{fig:swap} and \figref{fig:mag}.

Aas empirically verified in \cite{huang2022deep_amp},  
the degradation information in a weather image is mainly concentrated 
in the amplitude part of the Fourier domain. 
As shown in \figref{fig:swap}, when we swap the amplitude of the image with adverse weather conditions and its corresponding ground truth, we can see that the attributes of `degradation' and `clean` are transferred along with the amplitude. This validates our approach of utilizing amplitude to generate the degradation condition information. 

Furthermore, as shown in \figref{fig:mag}, the corresponding visualizations distinctly demonstrate how degradation is differently addressed for various weather patterns. For instance, images degraded by snow integrate higher frequency components due to snow particles. The snow removal process aligns these images closer to the lower frequencies seen in ground truth images. In contrast, haze removal increases the frequency details, as hazy images have lower frequency contents compared to their sharper ground truths. 
Therefore, we propose a degradation mapping loss to constrain the direction of degradation removal to align with the `clean' direction, thus further strengthening the degradation removal process.

For content reconstruction, As shown in Eqn.\ref{eq:fft}, each pixel in the Fourier space interacts with all the pixels in the spatial domain. We utilize this characteristic of Fourier to achieve global content reconstruction, which complements the local detail reconstruction. 
 
%-------------------------------------------------------------------------

\section{Experiments}

This section introduces our experimental setup and compares our performance with other techniques.

\begin{table}[t]
    \Large
    \centering
    \renewcommand\arraystretch{1.0}
    \caption{Quantitative results on the \textbf{[Setting1]} test dataset. PSNR / SSIM format. The best results are \textbf{boldfaced}.} 
    \vspace{-8pt}
    \resizebox{\linewidth}{!}{
    \begin{tabular}{c|c|cc|cc} 
    \toprule[1pt] 
    \multirow{2}{*}{Type}  & \multirow{2}{*}{Method} & \multicolumn{2}{c|}{Outdoor-Rain}  &\multicolumn{2}{c}{RainDrop}  \\  \cline{3-6}
     &  & PSNR $\uparrow$ & SSIM $\uparrow$ &PSNR $\uparrow$ &  SSIM $\uparrow$  \\    \hline
    \multirow{5}{*}{\centering Single-Task} & pix2pix \cite{isola2017image_pix2pix}                   &19.09   &0.71         &28.02  &0.85    \\
                                       & HRGAN \cite{li2019heavy_rain_hrgan}                      &21.56   &0.86          &  -    & -      \\  
                                       &Attn.Gan \cite{qian2018attentive_raind_raindrop_attngan}  & - & -        &30.55  &0.90    \\
                                       & MPRNet \cite{zamir2021multi_mprnet}            &21.90   &0.85            & - & -        \\
                                       & CCN \cite{quan2021removing_raind_ccn}          & -& -        &31.34  &0.95    \\  \hline
    \multirow{5}{*}{\centering Multi-Tasks} & All-in-One \cite{li2020all_all-in-one_all}     &24.71   &0.90        &31.12  &0.93    \\
                                            & TransWeather \cite{valanarasu2022transweather_all} &23.18   &0.84     &28.98  &0.90    \\
                                            & Chen \textit{et al.} \cite{chen2022learning_all}   &23.94   &0.85      &30.75  &0.91    \\ 
                                            & Zhu \textit{et al.} \cite{zhu2023learning_all}     &25.31   &0.90      &31.31  &0.93    \\                     
                                            & Ours                                          &\textbf{25.59}   & \textbf{0.90}               & \textbf{31.69}   & \textbf{0.93}             \\
    \bottomrule[1pt] 
    \end{tabular} }
    \label{tab:setting1}
\end{table}

\begin{table}[t]
\renewcommand\arraystretch{1.10}
\large
\vspace{-8pt}
\caption{Quantitative results on the \textbf{SnowTest100k-L} test dataset. PSNR / SSIM format. The best results are \textbf{boldfaced}.}
\resizebox{\linewidth}{!}{
\begin{tabular}{c|c|cc|cc} 
\toprule[1pt]
\multirow{2}{*}{Type}  &\multirow{2}{*}{Method}   &\multicolumn{2}{c|}{Setting1} &\multicolumn{2}{c}{Setting2} \\  \cline{3-6}
                      &                           & PSNR$\uparrow$ & SSIM$\uparrow$  & PSNR$\uparrow$  & SSIM$\uparrow$   \\ \hline

\multirow{4}{*}{\centering De-Snow} & DetailsNet \cite{deng2020detail_rain_drdnet}      &19.18   &0.75          &19.18   &0.75                                \\
& DesnowNet \cite{liu2018desnownet_snow100k}                                            &27.17   &0.90          &27.17   &0.90                                 \\
& JSTASR \cite{chen2020jstasr_snow}                                                     &25.32   &0.81          &25.32   &0.81                                 \\
& DDMSNET \cite{zhang2021deep_snow_ddmsnet}                                             &28.85   &0.88          &28.85   &0.88                                 \\  \hline
\multirow{4}{*}{\centering Multi-Tasks} & All-in-one \cite{li2020all_all-in-one_all}    &28.33   &0.88          & -      & -                                    \\
& Transformer \cite{valanarasu2022transweather_all}                                     &27.80   &0.85          &26.17   &0.88                                  \\
& Chen \textit{et al.} \cite{chen2022learning_all}                                      &29.27   &0.88          &28.71   &0.88                                   \\ 
& Zhu \textit{et al.} \cite{zhu2023learning_all}                                        &29.71   &0.89          &29.42   &0.89                                   \\
& Ours                                                                        &\textbf{30.59}   &\textbf{0.91} & \textbf{30.45}   & \textbf{0.90}                                  \\
\bottomrule[1pt]
\end{tabular}}
\label{tab:snow}
\end{table}

\begin{table}[t]
\Large
\renewcommand\arraystretch{1.1}
\setlength\tabcolsep{2pt}

\caption{Quantitative results on the \textbf{[Setting2]} test dataset. PSNR / SSIM format. The best results are \textbf{boldfaced}.}
\vspace{-10pt}
\resizebox{\linewidth}{!}{
\begin{tabular}{c|c|cc|c|cccl} 
\toprule[1pt] 
\multicolumn{1}{c|}{\multirow{2}{*}{Type}} &\multicolumn{3}{c|}{Rain14K}                                          &\multicolumn{3}{c}{RESIDE}        \\  \cline{2-7}
                                        & Method & PSNR$\uparrow$ & SSIM$\uparrow$                                           & Method & PSNR$\uparrow$ & SSIM$\uparrow$  \\    \hline
\multirow{7}{*}{\rotatebox{90}{Single-Task}} & JORDER \cite{yang2017deep_rain_jorder}              &31.28 &0.92          &EPDN \cite{qu2019enhanced_haze_epdn}          &23.82 & 0.89    \\
                                        & PReNet \cite{ren2019progressive_rain_prenet}        &31.88 & 0.93         &PFDN \cite{dong2020physics_haze_pfdn}         &31.45 & 0.97   \\
                                        & DRD-Net \cite{deng2020detail_rain_drdnet}           &29.65 & 0.88         &KDDN \cite{hong2020distilling_haze_kddn}      &33.49 & 0.97    \\
                                        & MSPFN \cite{dong2020multi_haze_msbdn}               &29.24 & 0.88         &MSBDN \cite{dong2020multi_haze_msbdn}         &33.79 & 0.98    \\
                                        & DualGCN \cite{fu2021rain_dualgcn}                   &30.50 & 0.91         &FFA-Net \cite{luo2021global_haze_ffanet}      &34.98 & 0.99   \\
                                        & JRJG \cite{ye2021closing_rain_jrjg}                 &31.18 & 0.91         &AECRNet \cite{wu2021contrastive_haze_aecrnet} &35.61 & 0.98   \\
                                        & MPRNet \cite{zamir2021multi_mprnet}                 &31.53 & 0.96         &MPRNet \cite{zamir2021multi_mprnet}           &31.31 & 0.97   \\  \hline
\multirow{5}{*}{\rotatebox{90}{Multi-Tasks}} & All-in-One \cite{li2020all_all-in-one_all}          &30.82 & 0.90         &All-in-One \cite{li2020all_all-in-one_all}    &30.49 & 0.95      \\
                                        & TransWeather \cite{valanarasu2022transweather_all}  &29.14 & 0.89         &TransWeather \cite{valanarasu2022transweather_all}    &27.66 & 0.95  \\
                                        & Chen \textit{et al.}\cite{chen2022learning_all}    &31.75 & 0.91         &Chen \textit{et al.} \cite{chen2022learning_all}              &30.76 & 0.97   \\
                                        & Zhu \textit{et al.}\cite{zhu2023learning_all}      &32.49 & 0.93         &Zhu \textit{et al.} \cite{zhu2023learning_all}                &30.85 & 0.98   \\
                                        & Ours                                               &\textbf{32.57} & \textbf{0.93}         & Ours     &\textbf{31.62} & \textbf{0.98}     \\
\bottomrule[1pt] 
\end{tabular}}
\vspace{-10pt}

  \label{tab:setting2}
\end{table}
\begin{figure*}[t]

\centering

\includegraphics[width=1.0\linewidth]{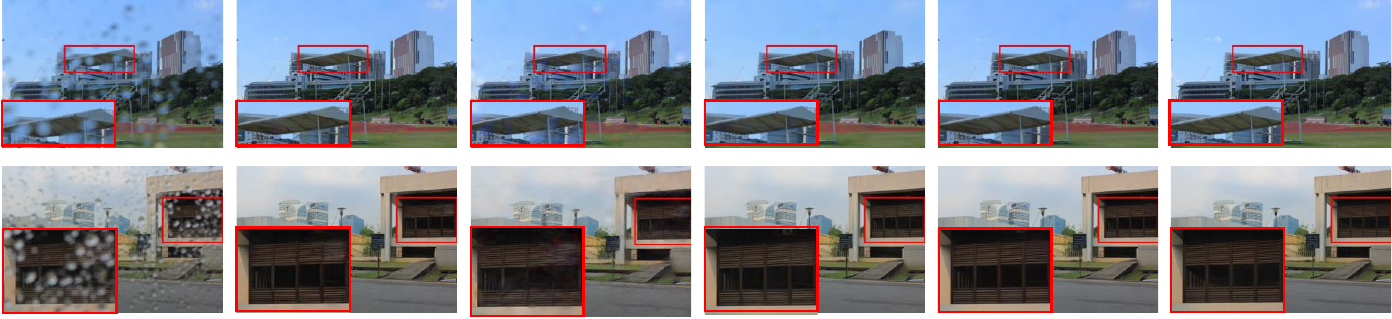} 
% \fbox{\rule{0pt}{3in} \rule{.9\linewidth}{0pt}}
\begin{tikzpicture}[overlay, remember picture]
  \node at (0, 0)[xshift=-\linewidth/(12/5)] {Input};  % 根据需要调整位置
  \node at (0, 0) [xshift=-\linewidth/4]{Chen \textit{et al.} \cite{chen2022learning_all}};
  \node at (0, 0)[xshift=-\linewidth/12] {TransWeather \cite{valanarasu2022transweather_all}};
  \node at (0, 0)[xshift=\linewidth/12] {Zhu \textit{et al.} \cite{zhu2023learning_all}};  % 根据需要调整位置
  \node at (0, 0) [xshift=\linewidth/4] {Ours};
  \node at (0, 0)[xshift=\linewidth/(12/5)] {Ground Truth};
\end{tikzpicture}
\vspace{-2pt}
\caption{Visual comparison of image deraining on the RainDrop dataset. Zoom in for better comparison.} 
\vspace{-4pt}

\label{fig:rain}
\end{figure*}
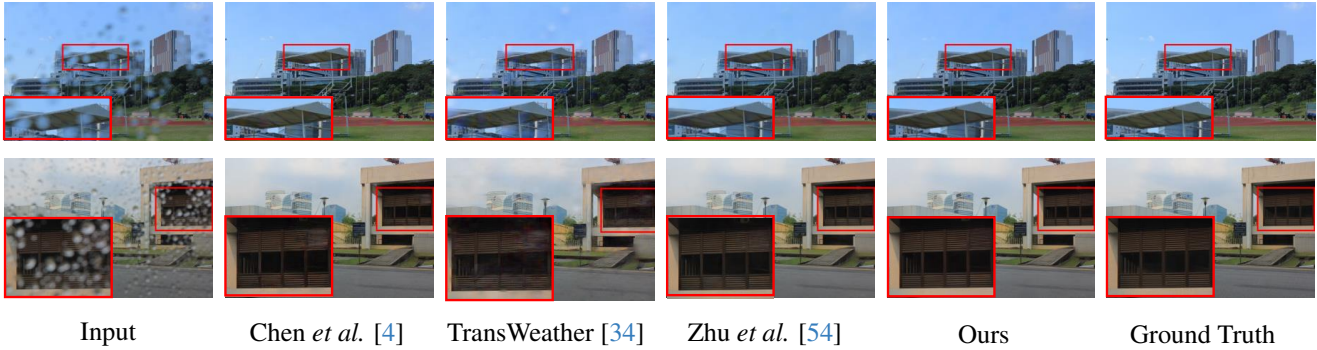

\begin{figure*}[t]

\centering

\includegraphics[width=1.0\linewidth]{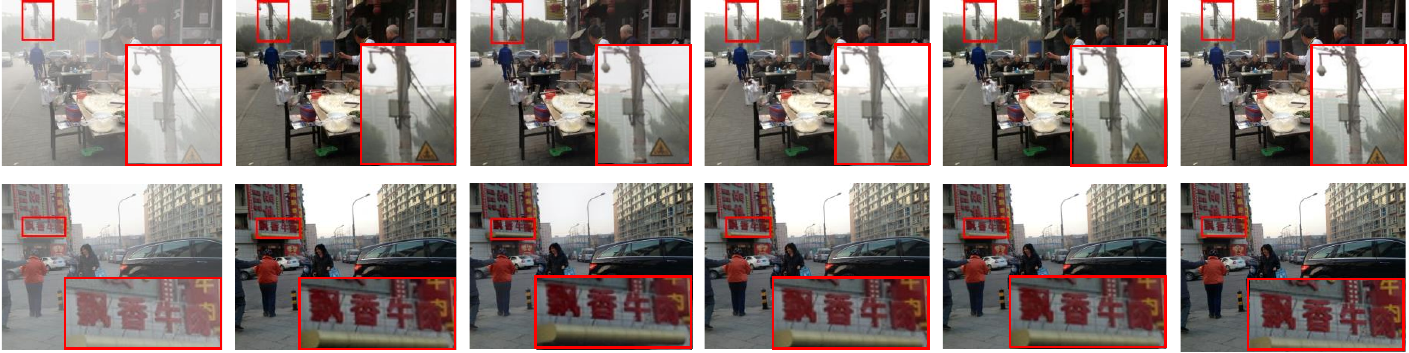} 
% \fbox{\rule{0pt}{3in} \rule{.9\linewidth}{0pt}}
\begin{tikzpicture}[overlay, remember picture]
  \node at (0, 0)[xshift=-\linewidth/(12/5)] {Input};  % 根据需要调整位置
  \node at (0, 0) [xshift=-\linewidth/4]{Chen \textit{et al.} \cite{chen2022learning_all}};
  \node at (0, 0)[xshift=-\linewidth/12] {TransWeather \cite{valanarasu2022transweather_all}};
  \node at (0, 0)[xshift=\linewidth/12] {Zhu \textit{et al.} \cite{zhu2023learning_all}};  % 根据需要调整位置
  \node at (0, 0) [xshift=\linewidth/4] {Ours};
  \node at (0, 0)[xshift=\linewidth/(12/5)] {Ground Truth};
\end{tikzpicture}
\vspace{-2pt}
\caption{Visual comparison of image dehazeing on the RESIDE dataset. Zoom in for better comparison.} 
\vspace{-2pt}

\label{fig:haze}
\end{figure*}

\subsection{Experimental Settings}

\paragraph{Datasets.}

To verify the effectiveness of our network framework, we employed two experimental dataset settings. 

\textbf{[Setting 1]} includes three different adverse weather datasets: `Raindrop' \cite{qian2018attentive_raind_raindrop_attngan}, `Outdoor-Rain' \cite{li2019heavy_rains_outdoor_rain}, and `Snow100K' \cite{liu2018desnownet_snow100k}. `Raindrop' consists of 861 paired training images and two test sets, Test A and Test B, which contain 58 and 249 paired images respectively. `Outdoor-Rain` comprises 9,000 paired training images and 1,500 paired testing images. `Snow100K` encompasses 50K paired training images and three test datasets: `Snow100K-S', `Snow100K-M', and `Snow100K-L', which contain 16,611, 16,588, and 16,801 paired images respectively. 
% Furthermore, it includes 1,329 real-world snow images under the category `realistic'. 
% To avoid long-tail distribution issues during training, we regulate the number of images in each training set to 9,000, adhering to the principle of 'reducing more and supplementing less.' For testing, we utilize Test A from the "Raindrop," all test sets from "Outdoor-Rain," and "Snow100K-L" from the "Snow100K" test sets. 

\textbf{[Setting 2]} also includes datasets of three different types of severe weather: `Rain14K' \cite{fu2017removing_rain_rain1400}, `RESIDE', and `Snow100K' \cite{liu2018desnownet_snow100k}, with the latter being the same as in Setting 1. `Rain14K' comprises 12,600 paired training images and 1,400 paired test images. `RESIDE' includes 313,944 paired training images and two test sets, `SOTS-Test-outdoor' and `SOTS-Test-indoor', each containing 500 paired images. 
% During training, we randomly sample 5,000 images from each training set. For testing, we use all images from "Rain14K," "SOTS-Test-indoor," and "Snow100K-L." 
To ensure fairness in comparisons, we employ the method described in \cite{valanarasu2022transweather_all, zhu2023learning_all} to uniformly sample images from the dataset during network training.

\paragraph{Implementation details.}
We conducted our experiments on four NVIDIA GTX 4090 GPUs. Each training iteration involves randomly shuffling images with a batch size of 32. 
Images are cropped to patches of size 224 $\times$ 224. Our network is trained for 300 epochs with the Adam optimization algorithm \cite{kingma2014adam}. The learning rate starts at 2e-4 and decays by cosine annealing scheme until 1e-6.

\subsection{Comparison with State-of-the-Art Methods}
To assess the performance of our network, we employed well-established evaluation metrics, including peak signal-to-noise ratio (PSNR) and structural similarity index (SSIM) \cite{SSIM}. These metrics are utilized 
as quantitative measures to evaluate the effectiveness of our network. 

We present our quantitative results in \tabref{tab:setting1}, \tabref{tab:snow}, and \tabref{tab:setting2}, including both all-weather-removal and typical-weather-removal methods. 
As can be seen from the tables, compared with all-weather-removal methods, 
our method achieved state-of-the-art performance across all synthetic test sets. We will present additional visualizations in the supplementary materials.
Furthermore, we demonstrate the visual effects of our approach in the \figref{fig:rain} and \figref{fig:haze}, which show that our method restores texture details more effectively, further illustrating the efficacy of our approach. Due to space constraints, we will present additional visual effect analysis in the supplementary materials. 

\begin{figure}[t]

\centering

\includegraphics[width=1.0\linewidth]{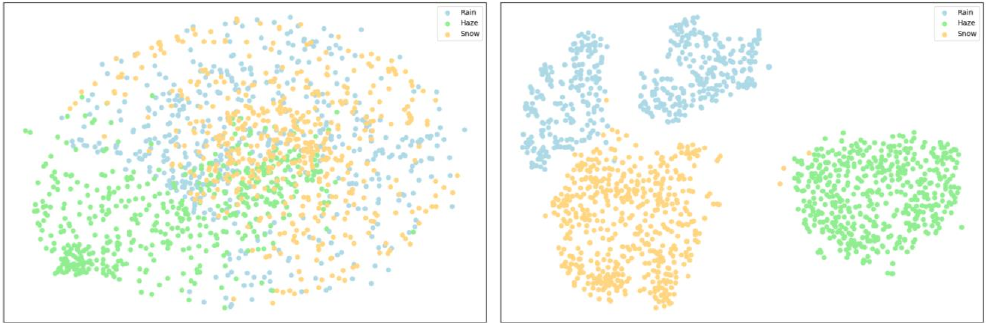} 
% \fbox{\rule{0pt}{3in} \rule{.9\linewidth}{0pt}}
\vspace{0pt}
\begin{tikzpicture}[overlay, remember picture]
  \node at (0, 0)[xshift=-0.25\linewidth] {(a)};  % 根据需要调整位置
  \node at (0, 0)[xshift=0.25\linewidth] {(b)};
\end{tikzpicture}
\caption{The t-SNE~\cite{van2008visualizing_tsne} visualization of the features from the last encoder block:
        (a) illustrates the input features of this block, and (b) depicts their amplitudes.} 
\vspace{-18pt}

\label{fig:tsne}
\end{figure}
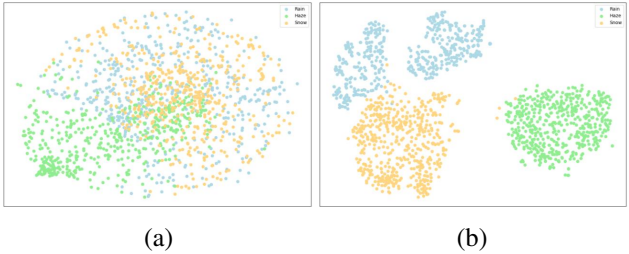

\subsection{Ablation Studies}
We conducted ablation experiments on the \textbf{[Setting 2]} Rain14K dataset 
to verify the effectiveness of our method.

\paragraph{Investigation of the DRM.}
To demonstrate the effectiveness of amplitude information from Fourier transforms in guiding the degeneration removal process, we omitted the amplitude guidance (AG) step from all degeneration removal modules in the network. This modified model is henceforth referred to as 'Model-1.' As shown in \tabref{tab:abl}, quantitative analysis confirms the usefulness of amplitude as a guide for degeneration removal. To further validate the effectiveness of extracting channel-independence degradation information, we removed the step involving subtraction of the norm amplitude (SNA) from the features, resulting in a variation named 'Model-2.' The quantitative results support the efficacy of this operation.

To more vividly demonstrate the rationale behind using amplitude as guidance information, we performed t-SNE \cite{van2008visualizing_tsne} visualization analysis on the features from the last encoder block. 
\figref{fig:tsne}(a) represents the input features of this block, while \figref{fig:tsne}(b) depicts their amplitudes. 
It is apparent that the amplitude can significantly distinguish different adverse weather types.

\paragraph{Investigation of the CRM.}
To demonstrate the efficacy of our proposed content restructuring module, we replaced all instances of this module within the network with convolutionally stacked RD Blocks, resulting in a modified architecture we refer to as 'Model-3.' Quantitative results, presented in \tabref{tab:abl}, indicate that our specifically designed module achieved superior performance.

\paragraph{Investigation of the Loss Function}
We also investigated the effect of different loss function combinations on network performance. We set $\lambda_2$ and $\lambda_3$ to 0 respectively to validate the effectiveness of these two loss functions. The quantitative results are shown in \tabref{tab:loss}, indicating that both loss functions contribute to performance improvement.

\paragraph{Parameter comparisons.}

We present a comparison of our method with others in terms of parameters and performance in \tabref{tab:param}, and it can be seen that our method achieves better performance with fewer parameters.

\begin{table}[t]
\small
\centering
\renewcommand\arraystretch{0.9}
\setlength\tabcolsep{3.0mm}
\vspace{-10pt}
\caption{Ablation study of our training strategy using the PNSR metric on the Rain14K of \textbf{[Setting 2]}. 
\textbf{AG}: omit the amplitude guidance step. \textbf{SNA}: remove subtraction of the norm amplitude.
\textbf{CR}: replaced all content reconstruction modules with RD Blocks.}
\vspace{-4pt}
\begin{tabular}{c|ccccccccccccl} 
\toprule 
\multirow {1}{*}{Method}    &AG               &SNA            & CR              & PSNR / SSIM   \\   \hline
Model-1                     &                 &               & \checkmark      & 32.24 / 0.93   \\                                      
Model-2                     & \checkmark      &               & \checkmark      & 32.42 / 0.93      \\
Model-3                     & \checkmark      & \checkmark    &                 & 32.39 / 0.93      \\
Ours                        & \checkmark      & \checkmark    & \checkmark      & 32.57 / 0.93     \\
\bottomrule 
\end{tabular} 
\label{tab:abl}
\end{table}
\begin{table}[t]
\small
\centering
\renewcommand\arraystretch{1.0}
\setlength\tabcolsep{5.2mm}
\vspace{-6pt}
\caption{Quantitative results(PSNR) of different loss function combinations.}
\vspace{-6pt}
% \resizebox{\linewidth}{!}{
\begin{tabular}{c|ccc|c} 
\toprule[1pt]
 No. & $\lambda_1$   & $\lambda_2$  & $\lambda_3$   & PSNR        \\   \hline
 1   & 1             & 1            & 0             & 32.37           \\ 
 2   & 1             & 0            & 1             & 32.34           \\ 
 3   & 1             & 1            & 1             & 32.57           \\ 
\bottomrule[1pt]

\end{tabular} 

  \label{tab:loss}
\end{table}
\begin{table}[t]
\small
\centering
\renewcommand\arraystretch{1.05}
\setlength\tabcolsep{0.6mm}
\vspace{-8pt}
\caption{The comparison between our method and other methods in terms of parameter and PSNR shows that our method has achieved the best overall performance.}
\vspace{-2pt}
% \resizebox{\linewidth}{!}{
\begin{tabular}{c|cccc} 
\toprule[1pt]
 Methods              & All-in-One  & TransWeather  & Chen~\textit{et al.}    & Ours        \\   \hline
 \#Param (M)         & 44.0         & 38.1          & 28.7                    & 11.2           \\ 
 PSNR(dB)            & 30.82        & 29.14        & 31.75                    & 32.57           \\ 
\bottomrule[1pt]
\end{tabular} 
\vspace{-6pt}
  \label{tab:param}
\end{table}

\section{Conclusion}

In this paper, we first identify the distinctions in degradation removal and content reconstruction processes for different adverse weather conditions. Then we decouple these two processes into distinct modules based on their channel statistics. The degradation removal module utilizes the observation that degradation information in images primarily concentrates on the Fourier domain's amplitude, generating  
channel-dependence degradation condition information to modulate the features along the channel dimension. This allows unique responses to different adverse weather types. The content reconstruction module mitigates the influence of degradation information by taking the mean of features along the channel dimension. Additionally, it exploits the global properties of the Fourier transform to enable consistent global content reconstruction. Our framework reduces the challenges of handling various adverse weather types in a unified network, thereby enhancing its performance.  The effectiveness of our method is demonstrated in various weather conditions.
{
    \small
    \bibliographystyle{ieeenat_fullname}
    \bibliography{main}
}

 \onecolumn
\clearpage
\setcounter{page}{1}
\maketitlesupplementary

\section{Channel Statistics}
\label{sec:rationale}
In the main text, \figref{fig:decon} warrants further explanation, which we provide here. Specifically, we delve deeper into the visualization of features, conducting a thorough examination of how different types of degradation manifest themselves as unique variations in the distribution of channels. This analysis not only illustrates the impact of degradation but also the nuances of its effects across various layers of the network.
Furthermore, we explore the role of the mean operation across channel dimensions in detail, emphasizing its effectiveness in diminishing the influence of degraded information. This process plays a crucial role in refining the quality of the content extracted from these images.

\begin{figure*}[h]
\centering
\includegraphics[height=4in, width=\linewidth]{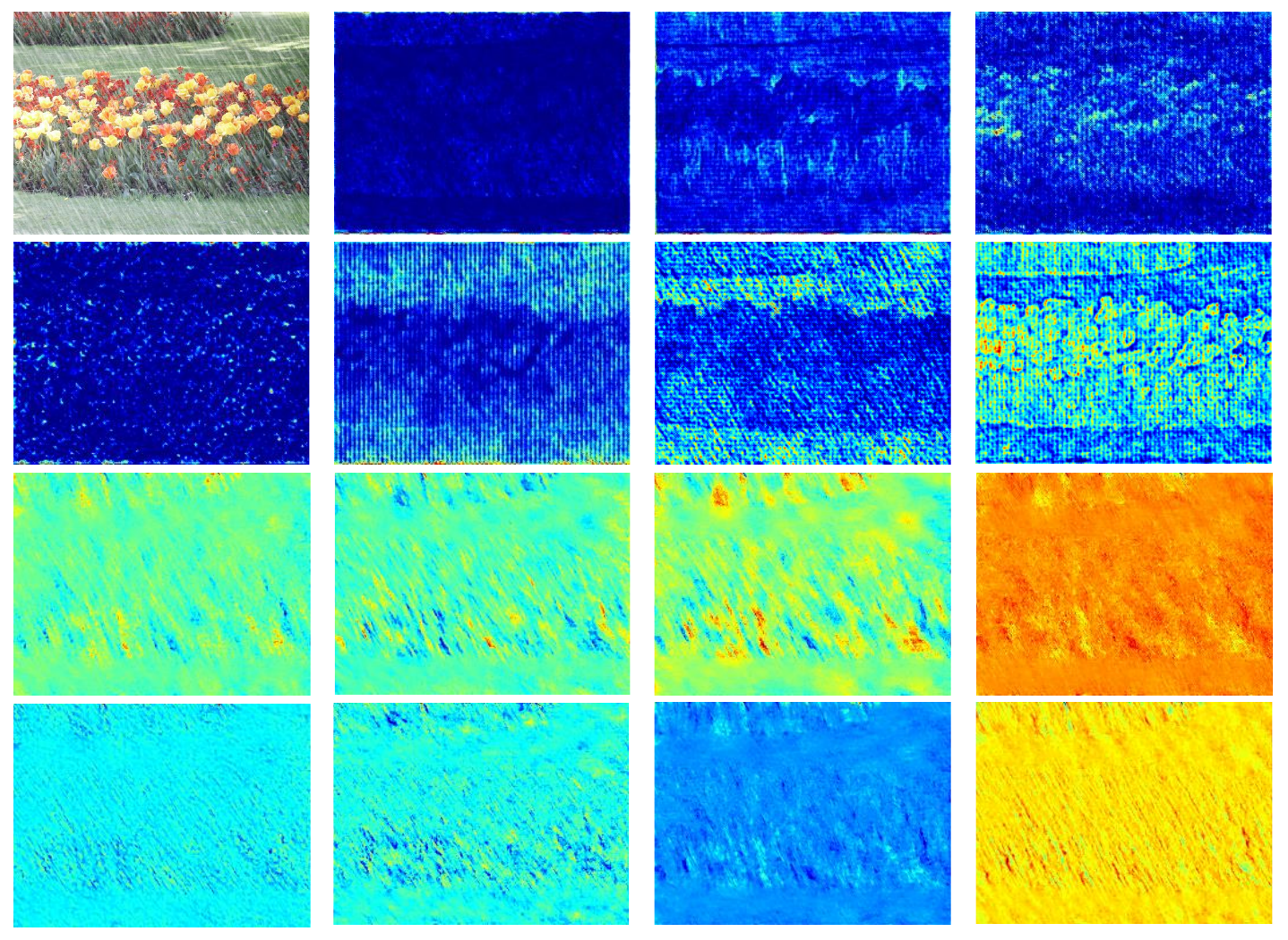}
\caption{Visualization of the intermediate features of \textbf{rain} images within the network (top two rows), and the corresponding gradients of the intermediate features (bottom two rows).}
\label{fig:supp_rain}
\end{figure*}

\begin{figure*}[h]
\centering
\includegraphics[height=4in, width=\linewidth]{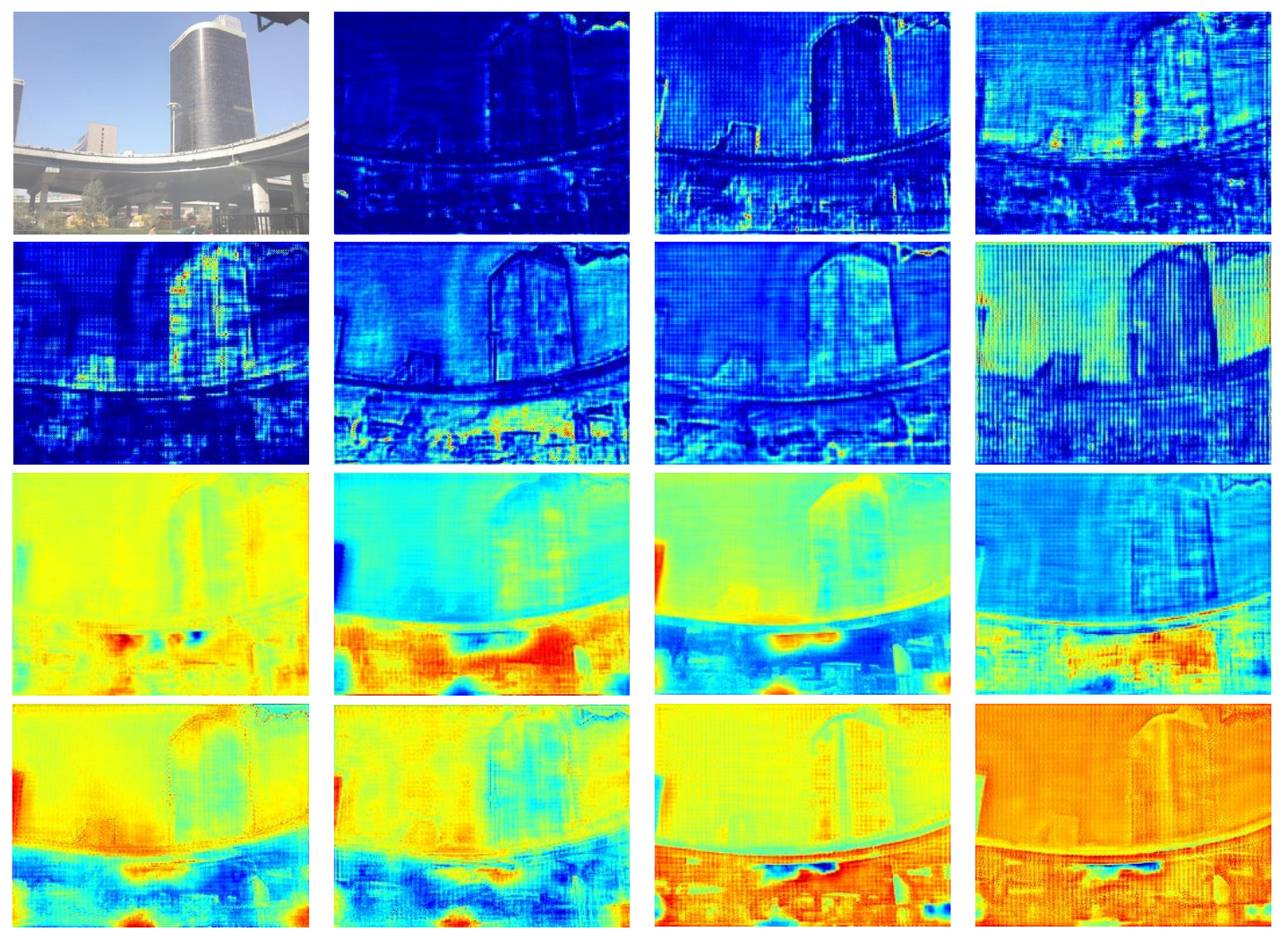}
\vspace{-10pt}
\caption{Visualization of the intermediate features of \textbf{haze} images within the network (top two rows), and the corresponding gradients of the intermediate features (bottom two rows).}
\vspace{-10pt}
\label{fig:supp_haze}
\end{figure*}

\begin{figure*}[h]
\centering
\includegraphics[height=4in, width=\linewidth]{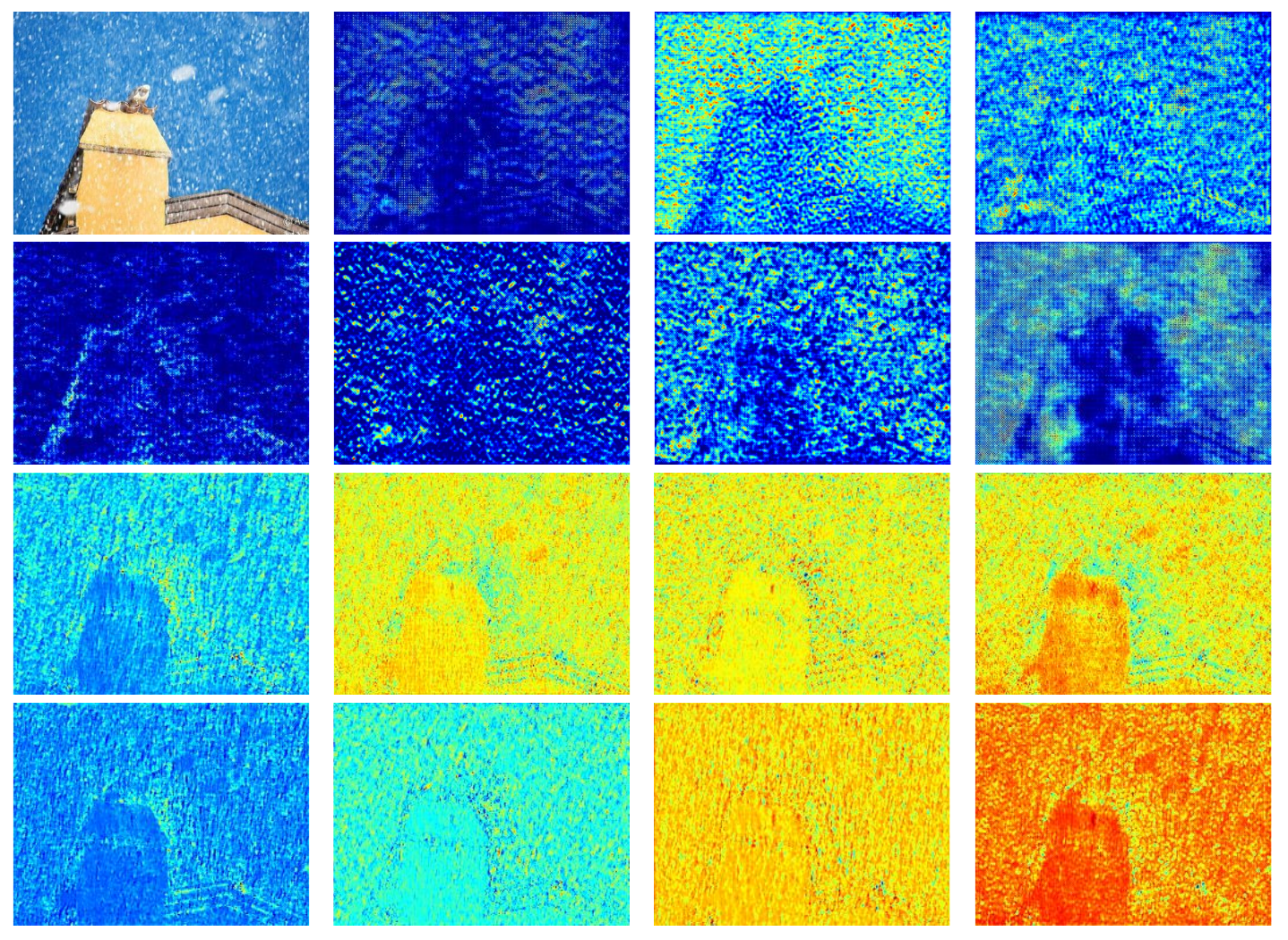}
\vspace{-10pt}
\caption{Visualization of the intermediate features of \textbf{snow} images within the network (top two rows), and the corresponding gradients of the intermediate features (bottom two rows).}
\vspace{-6pt}
\label{fig:supp_snow}
\end{figure*}

We have visualized images of various adverse weather conditions through the intermediate features and gradients at each feature layer within the network, as shown in \figref{fig:supp_rain}, \figref{fig:supp_haze} and \figref{fig:supp_snow}. This visualization process allows us to closely examine the intricate details and transformations that these images undergo during their progression through the network. By focusing on both the intermediate features and their corresponding gradients, we gain a comprehensive understanding of how different types of image degradation are processed and altered at each layer of the network.

\begin{figure*}[t]
\centering
\includegraphics[height=2.5in, width=\linewidth]{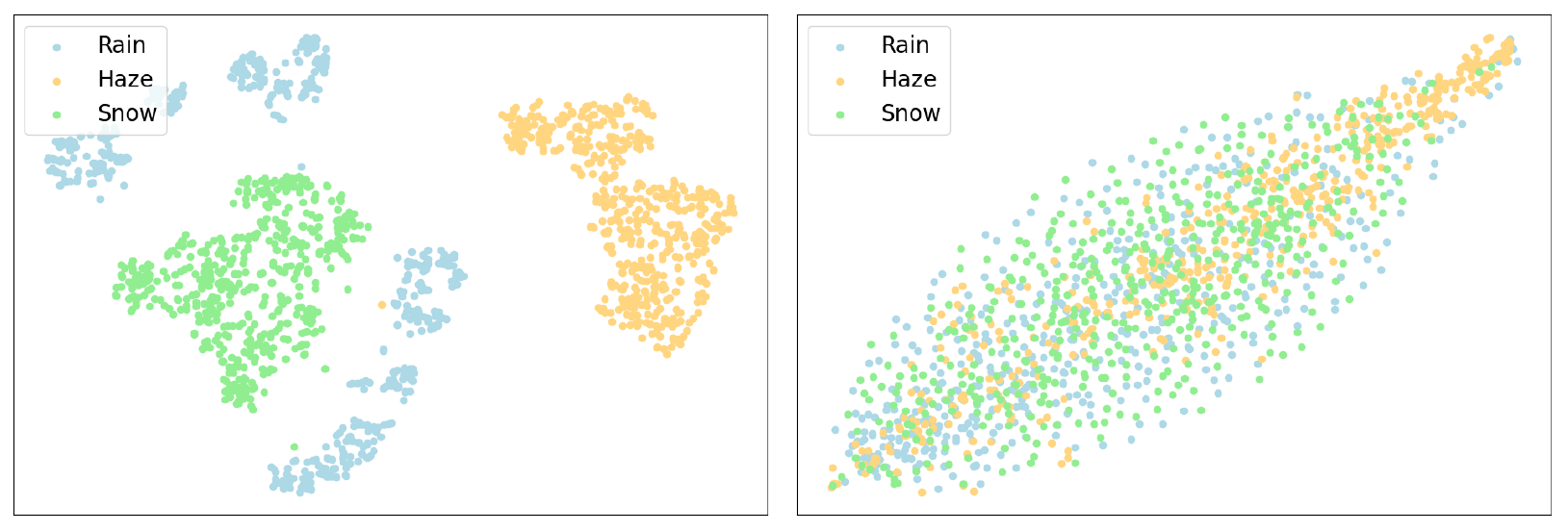}
\begin{tikzpicture}[overlay, remember picture]
  \node at (0, 0)[xshift=-0.25\linewidth] {(a)};  % 根据需要调整位置
  \node at (0, 0)[xshift=0.25\linewidth] {(b)};
\end{tikzpicture}
\caption{The t-SNE~\cite{van2008visualizing_tsne} visualization of the intermediate features, 
(a): before the mean operation across the channel dimension, (b): after the mean operation across the channel dimension.}
\label{fig:supp_tsne_mean}
\end{figure*}
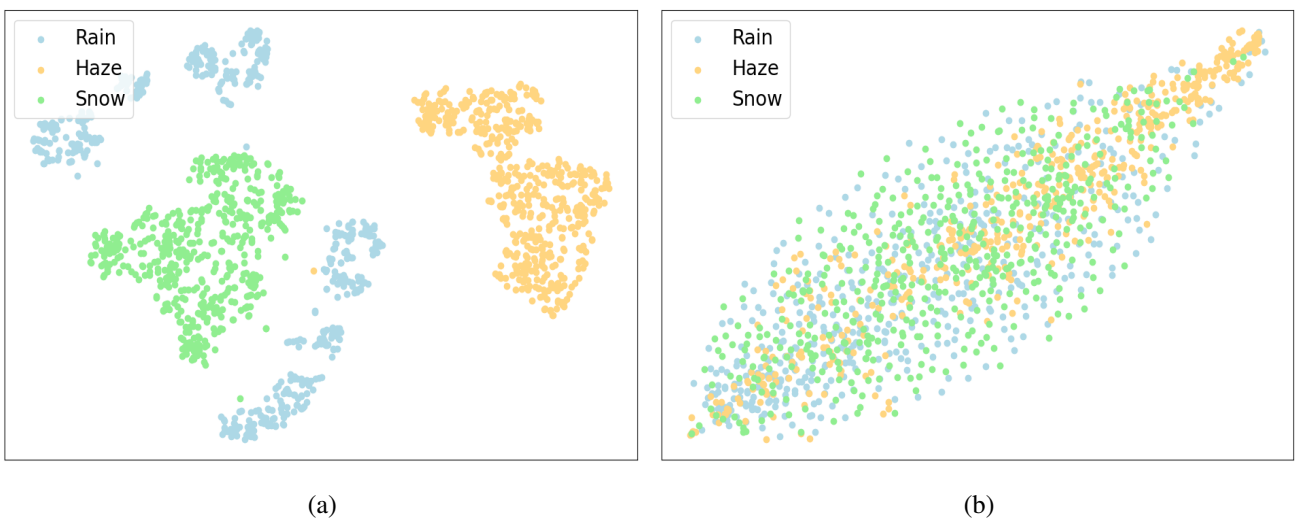

We input 500 images each of rain, haze, and snow into the network, and visualized the t-SNE clustering of their intermediate features. As shown in \figref{fig:supp_tsne_mean}, for instance, the left side illustrates how features are primarily clustered by the type of degradation before the mean operation across the channel dimension. However, on the right side, post-mean operation, the degradation information is markedly lessened. This reduction allows for a more content-centric clustering of features, highlighting the mean operation’s ability to enhance the network's focus on the essential aspects of the images, irrespective of the degradation type.


\begin{figure*}[t]
\centering
\includegraphics[width=\linewidth]{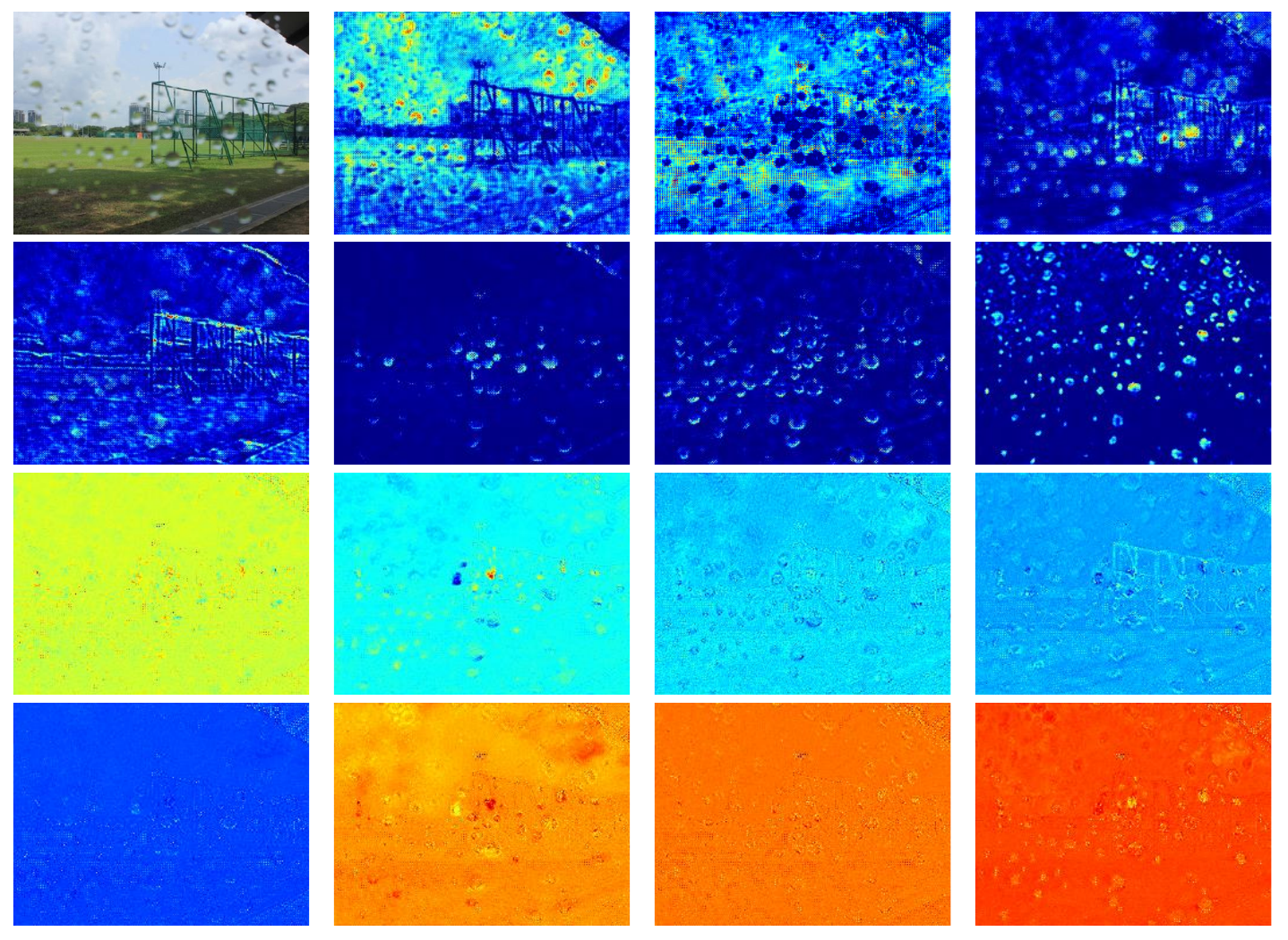}
\caption{Visualization of the intermediate features of \textbf{raindrop} images within the network (top two rows), and the corresponding gradients of the intermediate features (bottom two rows).}
\label{fig:supp_raindrop}
\end{figure*}

\clearpage

\begin{figure*}[t]

\centering

\includegraphics[height=3.5in, width=1.0\linewidth]{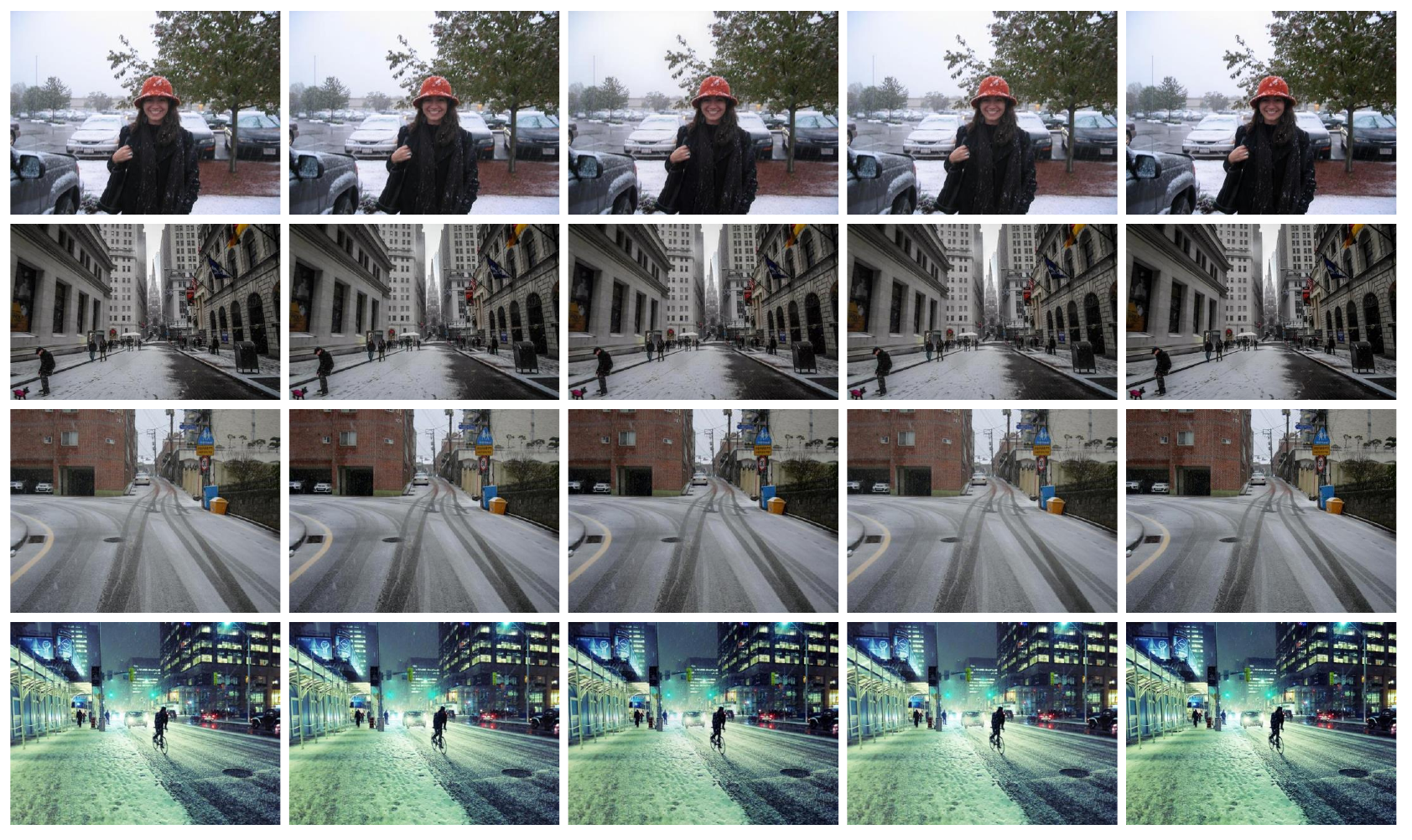} 
% \fbox{\rule{0pt}{3in} \rule{.9\linewidth}{0pt}}
\begin{tikzpicture}[overlay, remember picture]
  \node at (0, 0)[xshift=-\linewidth/(5/2)] {Input};  % 根据需要调整位置
  \node at (0, 0) [xshift=-\linewidth/5]{Chen \textit{et al.} \cite{chen2022learning_all}};
  \node at (0, 0)[xshift=0] {TransWeather \cite{valanarasu2022transweather_all}};
  \node at (0, 0) [xshift=\linewidth/5] {Zhu \textit{et al.} \cite{zhu2023learning_all}};
  \node at (0, 0)[xshift=\linewidth/(5/2)] {Ours};
\end{tikzpicture}
\vspace{-2pt}
\caption{Visual comparison of image desnowing on the realistic dataset. Zoom in for better comparison.} 
\vspace{-12pt}

\label{fig:real_snow}
\end{figure*}

\noindent % 确保没有缩进
\vspace{-6pt}
\begin{minipage}{0.5\textwidth}
    \centering
    \includegraphics[height=2in, width=\linewidth]{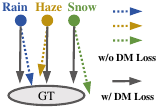}
    \captionof{figure}{The effect of the degradation mapping loss.}
    \label{fig:supp_dmloss}
\end{minipage}%
\begin{minipage}{0.5\textwidth}
    \centering
    \includegraphics[height=2in, width=\linewidth]{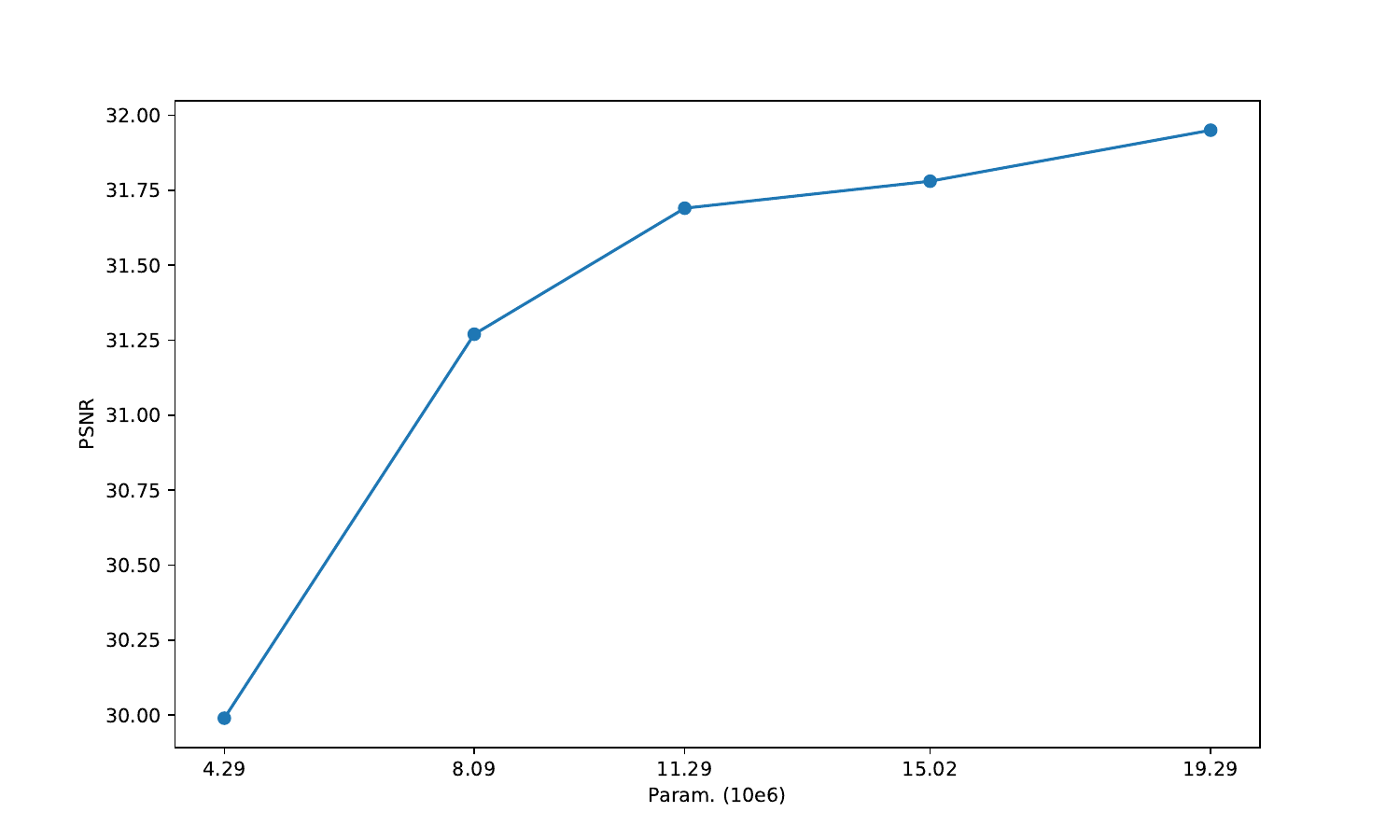}
    \captionof{figure}{Quantitative analysis of the network's performance under different parameter numbers.}
    \label{fig:supp_param}
\end{minipage}

\section{Experiments}
\paragraph{Snow visualation.}
Due to space limitations in the main text, we did not show the visual effect images of snow scenes. Here, we present a set of comparative visual effect images of snow scenes in \figref{fig:real_snow}. It can be seen that the images output by our network have better clarity and color contrast.

\paragraph{Degradation Mapping Loss.} 
We further provide a schematic illustration of the Degradation Mapping Loss constraint in \figref{fig:supp_dmloss}. It can be observed that without this loss, the optimization directions for different weather conditions vary. However, with the incorporation of this loss, the optimization directions for various weather conditions become consistent.

\paragraph{Ablation study of the parameter numbers.} 
We conduct a quantitative analysis on the test set of the Rain1400 dataset to evaluate the impact of different parameter quantities on network performance as shown in \figref{fig:supp_param}. It can be observed that with a smaller number of parameters, an increase in parameters leads to a significant improvement in performance. However, with a larger number of parameters, the enhancement in performance due to an increase in parameters is limited.

\end{document}